\documentclass[conference]{IEEEtran}
\IEEEoverridecommandlockouts

\usepackage{cite}
\usepackage{amsmath,amssymb,amsfonts}
\usepackage{algorithmic}
\usepackage{graphicx}
\usepackage{textcomp}
\usepackage{xcolor}
\def\BibTeX{{\rm B\kern-.05em{\sc i\kern-.025em b}\kern-.08em
    T\kern-.1667em\lower.7ex\hbox{E}\kern-.125emX}}

\usepackage[utf8]{inputenc}
\usepackage{algorithm}
\usepackage{epsfig}
\usepackage{mathtools}
\usepackage{multirow}
\usepackage{subcaption}
\usepackage{lipsum}
\usepackage{tablefootnote}
\usepackage{color}
\usepackage{dsfont}
\usepackage{booktabs}

\newcommand{\ie}{\textit{i.e.}}
\newcommand{\eg}{\textit{e.g.}}

\newcommand\blfootnote[1]{%
  \begingroup
  \renewcommand\thefootnote{}%
  \footnote{\noindent\hspace{-1em}\rule{0.2\textwidth}{0.4pt}\par #1}%
  \addtocounter{footnote}{-1}%
  \endgroup
}

\begin{document}

\title{Reward Generation via Large Vision-Language Model in Offline Reinforcement Learning}

\author{
\IEEEauthorblockN{Younghwan Lee$^\dagger$}
\IEEEauthorblockA{\textit{Electrical Engineering} \\
\textit{KAIST}\\
Daejeon, South Korea \\
youngh2@kaist.ac.kr}
\and
\IEEEauthorblockN{Tung M. Luu$^\dagger$}
\IEEEauthorblockA{\textit{Electrical Engineering} \\
\textit{KAIST}\\
Daejeon, South Korea \\
tungluu2203@kaist.ac.kr}
\and
\IEEEauthorblockN{Donghoon Lee}
\IEEEauthorblockA{\textit{Robotics Program} \\
\textit{KAIST}\\
Daejeon, South Korea \\
dh\_lee99@kaist.ac.kr}
\and
\IEEEauthorblockN{Chang D. Yoo}
\IEEEauthorblockA{\textit{Electrical Engineering} \\
\textit{KAIST}\\
Daejeon, South Korea \\
cd\_yoo@kaist.ac.kr}
}

\maketitle

\blfootnote{
$\dagger$ The authors are equally contributed.

This work was partly supported by Institute for Information \& communications Technology Planning \& Evaluation (IITP) grant funded by the Korea government(MSIT) (No.RS-2021-II211381, Development of Causal AI through Video Understanding and Reinforcement Learning, and Its Applications to Real Environments) and Institute of Information \& communications Technology Planning \& Evaluation (IITP) grant funded by the Korea government(MSIT) (No.RS-2022-II220184, Development and Study of AI Technologies to Inexpensively Conform to Evolving Policy on Ethics).}
\begin{abstract}

In offline reinforcement learning (RL), learning from fixed datasets presents a promising solution for domains where real-time interaction with the environment is expensive or risky. However, designing dense reward signals for offline dataset requires significant human effort and domain expertise. Reinforcement learning with human feedback (RLHF) has emerged as an alternative, but it remains costly due to the human-in-the-loop process, prompting interest in automated reward generation models. To address this, we propose Reward Generation via Large Vision-Language Models (RG-VLM), which leverages the reasoning capabilities of LVLMs to generate rewards from offline data without human involvement. RG-VLM improves generalization in long-horizon tasks and can be seamlessly integrated with the sparse reward signals to enhance task performance, demonstrating its potential as an auxiliary reward signal.
\end{abstract}


\section{Introduction}
Goal-conditioned offline reinforcement learning (RL) has emerged as a promising approach for applying deep RL in domains where real-time interaction with the environment is costly or impractical \cite{prudencio2023survey}, such as robotics \cite{singh2022reinforcement}, healthcare \cite{liu2020reinforcement}, and autonomous driving \cite{kiran2021deep}. Unlike online RL, where agents continuously interact with the environment, offline RL operates on fixed datasets, making it an attractive solution for scenarios where data collection is expensive or risky. To prevent distribution shifts between the offline dataset and the learned policy \cite{levine2020offline}, offline RL methods typically incorporate conservative learning objectives alongside value-based methods \cite{fujimoto2019off,kumar2020conservative,liu2020provably}. One of the central challenges in goal-conditioned offline RL is the sparsity of the reward signal \cite{andrychowicz2017hindsight,luu2021hindsight,ma2022offline,luu2024predictive}, where agents are unable to explore the environment to uncover informative states that align with the desired goals. Moreover, shaping a reward function to extract reward signals from the offline dataset may require extensive human effort and deep domain expertise \cite{zhu2023scaling,hadfield2017inverse}, and these handcrafted reward functions may still fail to fully reflect human intentions. 

Recently, reinforcement learning with human feedback (RLHF) has garnered increasing attention as a powerful paradigm that transforms human feedback into guidance signals, eliminating the need to manually design reward functions \cite{lee2021pebble, yuan2024uni, hiranaka2023primitive}. However, RLHF often requires a human-in-the-loop during the training of RL agents, which is costly. To address this, previous works have focused on learning a reward model to capture human preferences based on selected trajectories \cite{lee2021pebble, peng2020learning, xue2023reinforcement}. While promising, these methods still require human labeling on an initial dataset, which may lack flexibility when transferring to new tasks. Meanwhile, Large Vision-Language Models (LVLMs) trained on massive, Internet-scale datasets have demonstrated exceptional capabilities in solving complex reasoning problems \cite{alayrac2022flamingo,liu2024improved,reid2024gemini,achiam2023gpt}. Recent works have attempted to leverage LVLMs to tackle different aspects of the RL problem, such as generating planners \cite{ahn2022can,huang2022language}, discovering skills \cite{du2023guiding,zhang2023bootstrap}, or fine-tuning LVLMs for direct use as policies \cite{szot2023large,zhai2024fine}. 

\begin{figure}[t]
    \centering
    \includegraphics[width=0.7\linewidth]{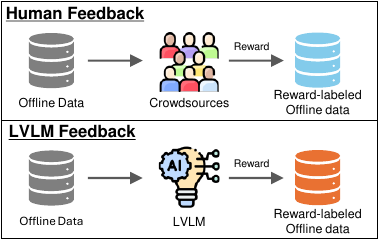}
    \caption{Unlike RLHF, we leverage the advanced reasoning capabilities of LVLM to generate rewards, eliminating the need for human involvement during the reward labeling process. This provides a more flexible and scalable approach that can be applied to a wide range of tasks.}
    \label{fig:compare_method}
    \vspace{-0.25in}
\end{figure}
\begin{figure*}[ht!]
    \centering
    \includegraphics[width=0.8\linewidth]{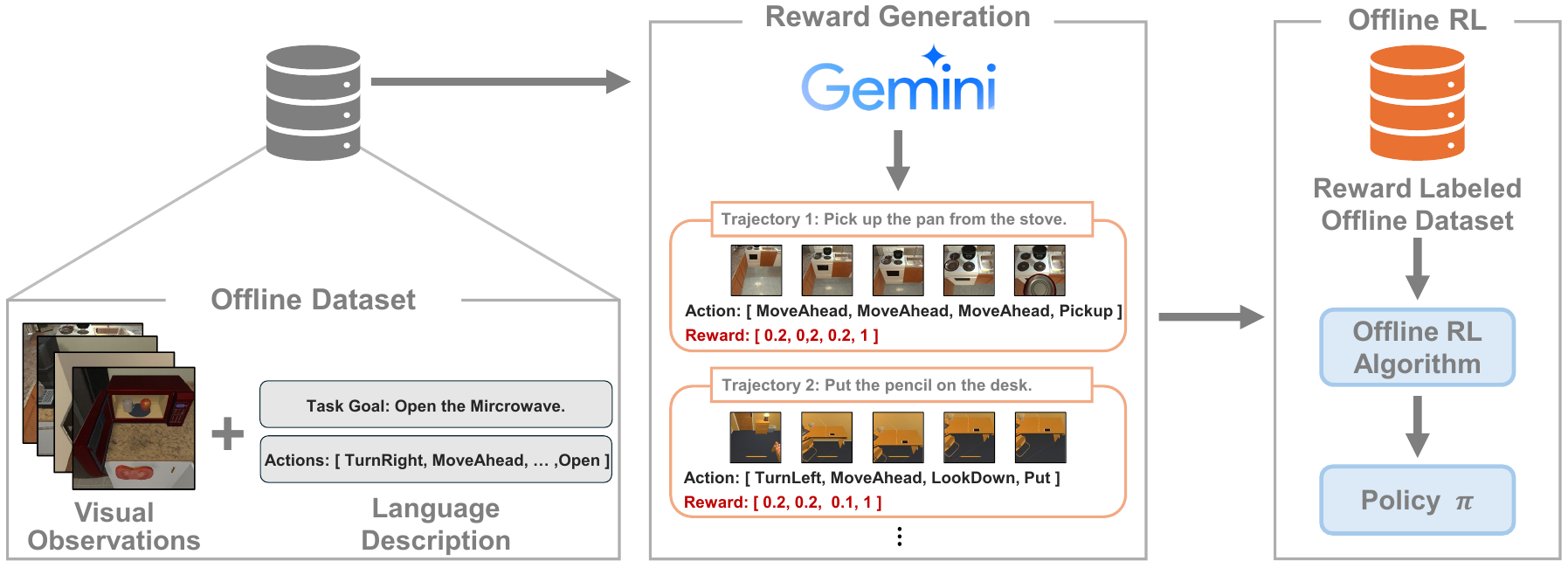}
    \caption{Algorithm Flow of RG-VLM and Offline RL training. In the RG-VLM reward generation process, the LVLM analyzes sequences of visual observations, actions, and task goals to generate rewards for the offline dataset. The reward labeled offline dataset is then used for offline RL training.}
    \label{fig:altorithm_flow}
    \vspace{-0.23in}
\end{figure*}

Motivated by these successes, and to address the challenge of approximating ground truth rewards in offline data, we propose \textbf{R}eward \textbf{G}eneration via L\textbf{VLM} (RG-VLM), a novel approach that leverages a pre-trained LVLM to generate rewards for previously collected data. In contrast to RLHF-based methods, RG-VLM eliminates the need for human involvement during the reward labeling process, as illustrated in Fig. \ref{fig:compare_method}, thereby offering a more scalable and flexible approach for generating rewards in new tasks. By harnessing the advanced reasoning capabilities of LVLM, RG-VLM generates interpretable dense rewards, providing more informative feedback and leading to improved performance in complex tasks.  To summarize, the key contributions of our work are:
\begin{enumerate}
    \item We introduce RG-VLM, a novel approach that leverages Large Vision-Language Models (LVLMs) to generate rewards from offline data, eliminating the need for human labeling and manual reward design.
    \item We demonstrate that RG-VLM enhances policy generalization across diverse long-horizon tasks, showing robust in varying environments and task complexities. 
    \item Our ablation study shows that combining RG-VLM with sparse rewards improves performance compared to using RG-VLM alone, highlighting its potential as an auxiliary reward signal to enhance traditional sparse rewards. 
\end{enumerate}

\section{Background}
\subsection{Problem Formulation}

We consider an RL agent that aims to complete $K$ target tasks $\{ \mathcal{T}_k \}_{1}^K \subset \mathcal{T}$, where $\mathcal{T}$ denotes the space of all tasks. For each task $\mathcal{T}_i \in \mathcal{T}$, the agent operates in a Markov decision process (MDP) defined as $\mathcal{M} = (\mathcal{S}, \mathcal{A}, P, R_i)$, where $\mathcal{S}$ is the state space (in our case, RGB images), $\mathcal{A}$ is the discrete action space, $P(s_{t+1}|s_t, a_t)$ is the transition probability in the robot environment, and $r_t^i = R_i(s_t, a_t, s_{t+1}) \in [0, 1]$ represents the shaped reward function, where $0$ indicates the task is not completed at all and $1$ indicates full completion of task $\mathcal{T}_i$. Lastly, we define $\mathcal{L}$ as the set of all natural language instructions, and let $\mathcal{L}_i \subset \mathcal{L}$ denote the set of language instructions that describe task $\mathcal{T}_i$. Note that there may be multiple instructions $l \in \mathcal{L}_i$ describing task $\mathcal{T}_i$, (\eg, ``pick up the butter knife from the countertop'' and ``grab the silver knife from the countertop''), and any particular instruction $l\in\mathcal{L}$ may describe multiple tasks (\eg, ``pick up the knife from the counter'' could refer to both picking up the butter knife and the yellow-handled knife). 

In this work, we assume that the true reward function $R_i$ for each task $\mathcal{T}_i$ is unobserved and must be inferred from natural language. Specifically, we assume access to an offline dataset $\mathcal{D}$ consisting of $N$ trajectories $\{\tau_j\}_1^N$, where each trajectory $\tau_j$ contains a sequence of states and actions, and a single linguistic instruction: $\tau_j=(\left[(s_0, a_0), (s_1, a_1), \dots, (s_T)\right], l_j)$. Furthermore, in our considered environment, each trajectory $\tau_j$ is assumed to be completed according to the instruction $l_j$; however, it is not necessarily an optimal trajectory (\eg, the trajectory might not be the shortest path to complete task $\mathcal{T}_i$). Note that the dataset $\mathcal{D}$ may contain tasks $\mathcal{T}_i$ that are not part of $\{ \mathcal{T}_k \}_{1}^K$. Our goal, then, is to infer rewards $\hat{r}_t^i = \hat{R}_i(s_t, a_t, s_{t+1})$ within the trajectory $\tau_j$ given the linguistic instruction $l$, states $s$, and actions $a$, where $\hat{R}$ is approximated by a prompted LVLM. Given trajectories relabeled with our reward function $\hat{R}$, we aim to learn a stochastic language-conditioned policy $\pi_{\theta}(a|s, l)$ to maximize the expected returns under the true reward function: $\mathbb{E}_{\pi}[\sum_{t=0}^T \gamma^t R_i(s_t, a_t, s_{t+1})]$ for the target task $\mathcal{T}_i \in \{ \mathcal{T}_k \}_{1}^K $, where $\gamma \in [0, 1)$ is a discount factor, and $T$ is episode horizon.

\subsection{Implicit Q-learning (IQL)}
One of the main challenges with offline RL is that a policy can exploit overestimated values for out-of-distribution actions \cite{levine2020offline}, as the agent cannot interact with the environment to correct erroneous policies and values, unlike in conventional online RL. In this work, we adopt IQL as an offline RL algorithm to learn the policy $\pi_{\theta}$, which avoids querying out-of-distribution actions by replacing the $\max$ operator in the Bellman optimality equation with expectile regression. We also extend IQL to condition on linguistic instructions. Specifically, IQL learns an action-value function $Q_{\omega}(s, a, l)$ and a state-value function $V_{\phi}(s, l)$ by optimizing the following objectives:
\begin{equation} \label{eq:v_loss}
    \small
    J_V(\phi) = \mathbb{E}_{(s,a,l)\sim\mathcal{D}}\left[L_2^{q}(Q_{\bar{\omega}}(s,a,l) - V_{\phi}(s,l))\right]
\end{equation}
\begin{equation}\label{eq:q_loss}
    \small
    J_Q(\omega) = \mathbb{E}_{(s,a,s', \hat{r}, l)\sim \mathcal{D}} \left[(\hat{r} + \gamma V_{\phi}(s',l) - Q_{\omega}(s, a, l))^2\right]
\end{equation}
where, $\hat{r}$ is our generated reward, $\bar{\omega}$ denotes the parameters of the target $Q$ network, and $L_2^q$ is the expectile loss with a parameter $q\in[0.5, 1): L_2^q = |q-\mathds{1}(x < 0)|x^2$. After training the value functions with Eq. (\ref{eq:v_loss}) and (\ref{eq:q_loss}), IQL learns the policy with advantage-weighted regression (AWR) \cite{ashvin2020accelerating}:
\begin{equation}
    \small
    J_{\pi}(\theta) = \mathbb{E}_{(s, a, s', l)\sim\mathcal{D}} [e^{\beta(Q_{\bar{\omega}}(s, a, l) - V_{\phi}(s, l))}\log \pi_{\theta}(a|s, l)]
\end{equation}
where, $\beta\in\mathbb{R}^{+}$ denotes an inverse temperature parameter.

\begin{figure}[t]
    \centering
    \includegraphics[width=0.7\linewidth]{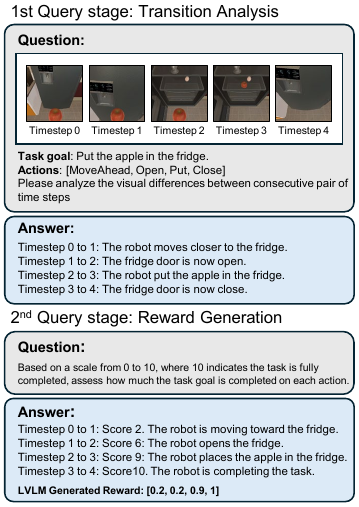}
    \caption{Example of RG-VLM querying process. The LVLM is queried in two stages. In the first stage, the LVLM analyzes sequences of visual observations, actions, and the task goal to understand how actions affect task progression. In the second stage, the LVLM assigns reward scores (0 to 10) based on each action's contribution to the task.}
    
    \label{fig:prompt_example}
    \vspace{-0.23in}
\end{figure}

\section{Proposed Method}
In this section, we introduce RG-VLM, a method designed to label offline data with rewards via Large Vision-Language Models, such as Gemini 1.5 pro \cite{reid2024gemini}, GPT-4,\cite{achiam2023gpt}, and Flamingo\cite{alayrac2022flamingo}. As shown in Fig. \ref{fig:altorithm_flow}, RG-VLM takes sequences of visual observations, corresponding actions, and the task goal, \ie, linguistic instruction, from the offline dataset as input to the LVLM. The LVLM analyzes each action, scoring it based on its contribution to the task goal. This process generates rewards for each transition, resulting in reward labeled offline dataset. The reward-labeled offline dataset is then used to learn the policy with any offline RL algorithms.

The querying process in RG-VLM is performed in two stages to generate rewards from offline data. In the first stage, a sequence of visual observations is concatenated into a single image, and the corresponding actions and task goal are provided as text. The LVLM analyzes the differences between the observations, enabling it to understand the transitions and assess how each action contributes to achieving the goal. In the second stage, the LVLM is queried again to assign a reward to each action, scoring them on a scale from 0 to 10, with 10 representing full contribution to the goal. An example of querying process is shown in Fig. \ref{fig:prompt_example}. This two-turn querying process allows the model to first interpret the context of the transitions, leading to more accurate and interpretable reward assignments in the second stage. 

To make this process is efficient, we fix the window size to 8. For each window, 8 transitions are collected and their corresponding visual observations are concatenated into a single image. If the window size is too small, a large number of queries would be required to cover the entire trajectory, significantly increasing computational cost and time. Conversely, a large window size could result in the concatenated image that is either too large or loses important detail when resized, making it difficult for the LVLM to analyze the transitions accurately. Through preliminary experiments, we found that a window size of 8 strikes the optimal balance, preserving image quality while minimizing the number of queries needed for each trajectory.

The reasoning capability of the LVLM enables RG-VLM to generate interpretable rewards by understanding the context of both visual observations and actions. During the second query, the LVLM assigns low or zero rewards to actions that are not relevant to the task goal, while accurately rewarding actions that significantly contribute to task progression. This process leverages the LVLM's strong reasoning abilities, allowing for more precise and meaningful reward assignments.

\begin{algorithm}[t]
\small
\caption{RG-VLM: LVLM Reward Generation and Offline RL Training}\label{alg:dr_vlm} 
\begin{algorithmic}[1]
    \STATE \textbf{Input:} A pre-trained LVLM, offline dataset $\mathcal{D}$, window size $w$

    \STATE \textbf{Initialize:} The policy $\pi_{\theta}$
    \STATE \texttt{//RG-VLM: Reward Generation via LVLM}
    \FOR{each trajectory $\tau_i \in \mathcal{D}$}
         \FOR{$t = 1$ to total length $T_{\tau_i}$ with step $w$}

            \STATE Extract samples $\{(s_j, a_j, s_{j+1})_{j=t}^{t+w-1}\}$
            \STATE Concatenate $\{s_j\}_{j=t}^{t+w}$ into an image $\hat{s}$
            \STATE Formulate prompt with $\hat{s}$, $\{a_j\}_{j=t}^{t+w-1}$ and goal $g$
            \STATE Query LVLM to analyze transitions 
            \STATE Query LVLM again to get rewards $\{\hat{r}_j\}_{j=t}^{t+w-1}$

            \STATE Store $\{\hat{r}_j\}_{j=t}^{t+w-1}$ into $\mathcal{D}$
        \ENDFOR
    \ENDFOR

    \STATE \texttt{//Offline RL Policy Training}
    \WHILE{not converged}
        \STATE Sample relabeled trajectories from $\mathcal{D}$
        \STATE Train the policy $\pi$ using an offline RL algorithm 
    \ENDWHILE
\end{algorithmic}
\end{algorithm}
\setlength{\textfloatsep}{0.1in}

\begin{figure*}[t]
    \centering
    \includegraphics[width=0.8\linewidth]{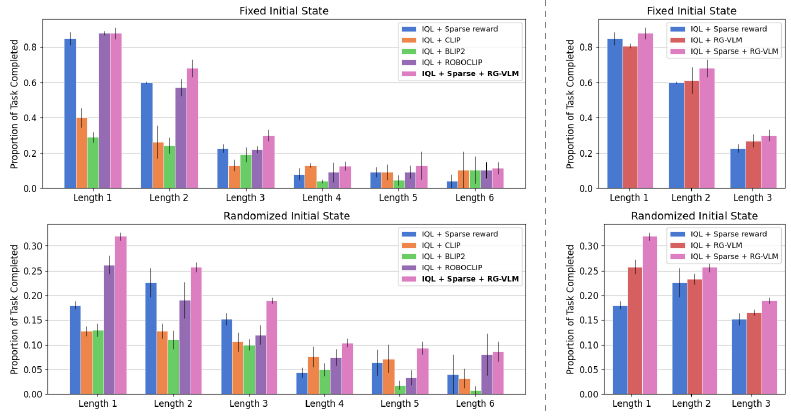}
    \caption{Proportion of Task Completion for 5 different methods across tasks with 1 to 6 sub-tasks. Left: Comparison of IQL with Sparse and RG-VLM rewards against other VLM-based methods and Sparse rewards only. Right: Ablation study showing the impact of combining RG-VLM with sparse rewards across various task lengths.}
    \label{fig:exp_result}
    \vspace{-0.23in}
\end{figure*}

\section{Experiments}
\subsection{Experimental setup}
\textbf{Offline Dataset. } To evaluate the effectiveness of RG-VLM, we use the customized dataset of ALFRED \cite{zhang2024sprint}, which processes the redundant instructions in the original dataset \cite{shridhar2020alfred}. Our training involved $73k$ language-annotated skill trajectories, with each method trained using five different seeds. Evaluation was conducted on 100 tasks of varying lengths from 1 to 6 sub-tasks. We further evaluated the trained agents in two initialization settings: (1) a fixed initial state, where the agent's initial state matches the one from the offline dataset, and (2) a randomized initial state, where the agent's position is randomly initialized, differing from the offline dataset. During evaluation, the agent received a reward of 1 only upon successfully completing the task goal. The return was measured as the proportion of the task successfully completed.

\textbf{Evaluated Methods. } We compare our method against four baselines for reward labeling: (1) IQL with sparse rewards, where we assign a reward of 1 to the final state in each trajectory; (2) VLM-RMs\cite{rocamonde2023vision} (IQL with CLIP \cite{radford2021learning}) ; and (3) IQL with BLIP2 \cite{li2022blip}, both of which assign rewards based on the similarity score between the task goal and a single observation image; (4) IQL with RoboCLIP \cite{sontakke2024roboclip}, which assigns rewards based on a sequence of images and the task goal. Our method, labeled as IQL with Sparse and RG-VLM Reward, combines RG-VLM rewards for all actions and assigns a sparse reward to the final action. For RG-VLM, we employed Gemini-1.5 pro \cite{reid2024gemini} as the LVLM in this experiment.

\subsection{Experimental Results}
Figure \ref{fig:exp_result} presents the return values of 5 different methods across tasks of varying lengths, ranging from 1 to 6 sub-tasks. Among all methods, IQL with Sparse and RG-VLM reward demonstrated superior performance, achieving the highest return across both fixed and randomized initial state conditions. We assessed generalization ability by comparing performance drops due to randomized initial positions. IQL with Sparse reward and IQL with RoboCLIP showed a performance decrease of 49.91\% and 47.74\%. In contrast, IQL with Sparse and RG-VLM reward exhibited only a 38.94\% reduction, highlighting its stronger generalization to unseen conditions.

Notably, IQL with Sparse and RG-VLM reward achieved 1.12x - 3.15x higher return against other VLM-based methods under fixed initial state condition. This result emphasizes the reasoning ability of Large Vision-Language Models (LVLMs) in generating rewards, leading to better task performance.

\subsection{Ablation Study}
As shown on the right side of Fig.\ref{fig:exp_result}, our ablation study indicates that combining RG-VLM rewards with sparse rewards achieved 1.14x higher returns on average, demonstrating the effectiveness of RG-VLM as an auxiliary reward signal. Additionally, Gemini 1.5 Pro achieved the best balance of task completion accuracy and inference time for reward generation in large offline datasets, as summarized in Table \ref{tab:accuracy_tradeoff}.

\begin{table}[h!]
\small
\centering
\begin{tabular}{lcc}
\toprule
Model               & Accuracy (\%) & Inference Time (s) \\
\midrule
Gemini 1.5 Pro      & 93.4          & 26.6               \\
Gemini 1.5 Flash    & 74.2          & 17.0               \\
Qwen2-VL-72B        & 86.2          & 132.8              \\
\bottomrule
\end{tabular}

\caption{Task completion accuracy and Inference time on 500 random trajectories with different LVLMs. Gemini models are called via public API and Qwen2-VL-72B is run on a single A100 GPU.}
\label{tab:accuracy_tradeoff}
\vspace{-0.2in}
\end{table}

\section{Conclusion and Future Work}
In this paper, we introduced RG-VLM, a novel method that leverages Large Vision-Language Models (LVLMs) to generate reward labels from offline data, eliminating the need for human intervention. By generating more informative reward signals, RG-VLM improves policy generalization and performance in long-horizon tasks. Additionally, we showed that RG-VLM can complement sparse reward structures. Future work will explore extending RG-VLM to more complex environments and tasks, such as Calvin\cite{mees2022calvin}, and integrating more advanced reasoning capabilities of LVLMs to improve reward precision.

\bibliographystyle{IEEEtran}
\bibliography{refs}

\end{document}